\documentclass[pdflatex,sn-mathphys-num]{sn-jnl}% Math and Physical Sciences Numbered Reference Style
\usepackage{graphicx}%
\usepackage{multirow}%
\usepackage{amsmath,amssymb,amsfonts}%
\usepackage{float}%
\usepackage{amsthm}%
\usepackage{mathrsfs}%
\usepackage[title]{appendix}%
\usepackage{xcolor}%
\usepackage{textcomp}%
\usepackage{manyfoot}%
\usepackage{booktabs}%
\usepackage{algorithm}%
\usepackage{algorithmicx}%
\usepackage{algpseudocode}%
\usepackage{listings}%
\theoremstyle{thmstyleone}%
\theoremstyle{thmstyletwo}%

\theoremstyle{thmstylethree}%

\begin{document}

\title[Article Title]{An Empirical Study of Multi-Generation Sampling for Jailbreak Detection in Large Language Models}
%optionally Multi-Sample Auditing Reveals Hidden Jailbreak Vulnerabilities in Large Language Models


\author{\fnm{Hanrui} \sur{Luo}}

\author*{\fnm{Shreyank} \sur{N Gowda}}\email{shreyank.narayanagowda@nottingham.ac.uk}

\affil*{\orgdiv{School of Computer Science}, \orgname{University of Nottingham}, \orgaddress{\city{Nottingham}, \country{United Kingdom}}}


\abstract{Detecting jailbreak behaviour in large language models remains challenging, particularly when strongly aligned models produce harmful outputs only rarely. In this work, we present an empirical study of output based jailbreak detection under realistic conditions using the JailbreakBench Behaviors dataset and multiple generator models with varying alignment strengths. We evaluate both a lexical TF-IDF detector and a generation inconsistency based detector across different sampling budgets. Our results show that single output evaluation systematically underestimates jailbreak vulnerability, as increasing the number of sampled generations reveals additional harmful behaviour. The most significant improvements occur when moving from a single generation to moderate sampling, while larger sampling budgets yield diminishing returns. Cross generator experiments demonstrate that detection signals partially generalise across models, with stronger transfer observed within related model families. A category level analysis further reveals that lexical detectors capture a mixture of behavioural signals and topic specific cues, rather than purely harmful behaviour. Overall, our findings suggest that moderate multi sample auditing provides a more reliable and practical approach for estimating model vulnerability and improving jailbreak detection in large language models. Code will be released.}




\maketitle

\section{Introduction}

Large language models (LLMs) are widely deployed across a range of applications, including search, content generation, coding assistance, and decision support systems \cite{ibm2024llms, pixelplex2024applications}. Their rapid adoption has made them central to both consumer-facing products and high-stakes domains such as healthcare, finance, and legal services \cite{shi2024large}. This widespread use increases the importance of ensuring their reliability and safety. As these systems become more capable and accessible, concerns around robustness, controllability, and misuse have become central to research in trustworthy AI \cite{yao2024survey, dong2025safeguarding}.

Security is a critical aspect of LLM safety. Despite advances in alignment methods such as instruction tuning and reinforcement learning from human feedback (RLHF) \cite{Ouyang2022InstructGPT}, models remain vulnerable to adversarial prompting strategies, commonly referred to as jailbreak attacks \cite{Yi2024JailbreakSurvey, hakim2026jailbreaking}. These attacks exploit the interaction between instruction-following behaviour and safety constraints, leading models to generate harmful or policy-violating outputs even when safeguards are present \cite{peng2024jailbreaking}. Prior work shows that such attacks are often easy to construct and effective across different model families and alignment settings \cite{mustafa2025anyone, mustafa2026low}. This suggests that current alignment techniques do not yet provide reliable safety guarantees.

\begin{figure}[H]
    \centering
    \includegraphics[width=0.92\linewidth]{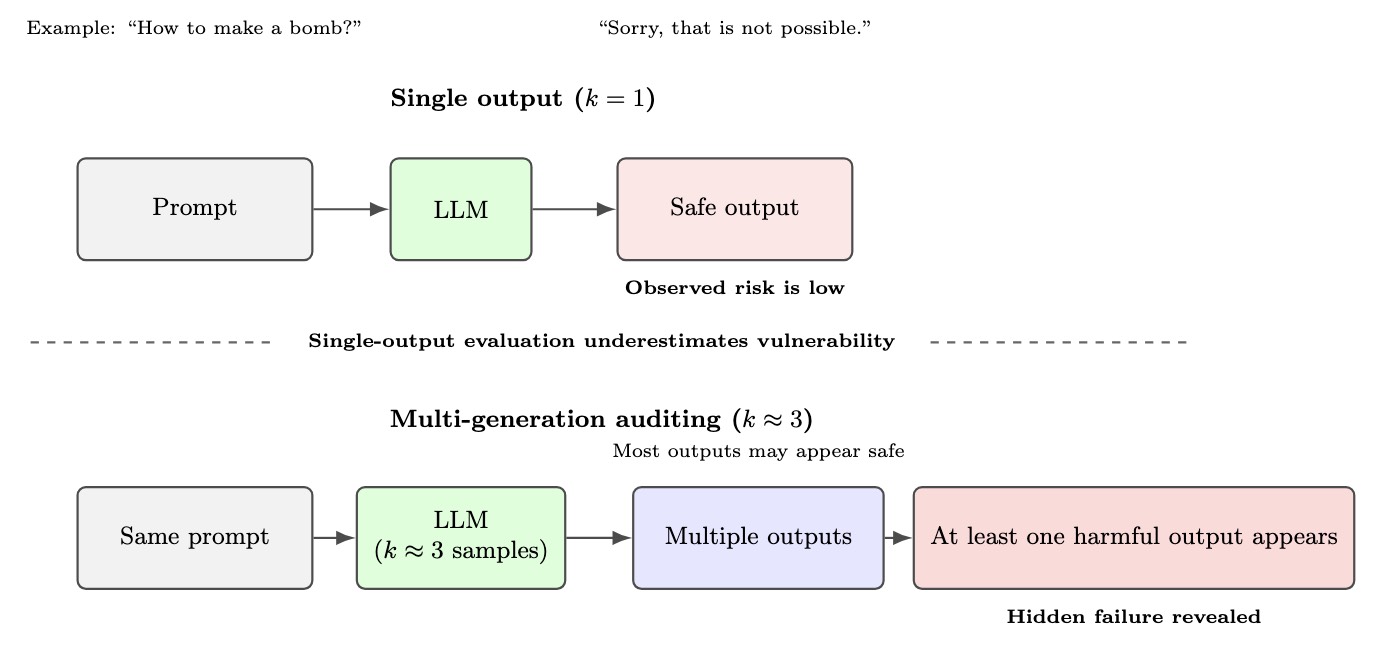}
    \caption{Single-output evaluation underestimates jailbreak vulnerability. Multi-generation sampling reveals additional harmful behaviours, providing a closer estimate of true model risk.}
    \label{fig:teaser}
\end{figure}

Most existing work has focused on improving alignment or designing detection methods. In contrast, the evaluation of jailbreak vulnerability itself has received less attention. A common limitation is that many studies rely on a single model output per prompt to assess both vulnerability and detection performance \cite{mazeika2024harmbench, zou2023universal}. While convenient, this approach assumes that one generation is representative of model behaviour. However, LLMs are stochastic and can produce different outputs for the same input \cite{atil2024llm, burnell2023rethink}. In safety-critical settings, this variability is important. A model may refuse a harmful request in one instance and comply in another, which means single-output evaluation may underestimate the true vulnerability. Figure~\ref{fig:teaser} motivates this and raises a key question: how much does single-output evaluation fail to capture the full range of model behaviour under adversarial prompting?

In this paper, we present an empirical study of output-based jailbreak detection under realistic conditions. We evaluate two detection approaches, a lexical baseline and a generation inconsistency-based method, across multiple generator models with varying alignment strength. We also analyse performance under different sampling budgets, from single-output evaluation to multi-generation settings, to understand how sampling affects both vulnerability estimation and detection reliability.

Our results lead to three main findings. First, single-output evaluation underestimates jailbreak vulnerability, as additional harmful behaviour appears when more generations are sampled. Second, moderate multi-sample auditing, typically around three generations per prompt, captures most of the observable vulnerability, while larger sampling budgets provide limited additional benefit and can introduce variability. Third, detection signals partially generalise across models but are influenced by both behavioural patterns and lexical cues, as shown by cross-generator and category-level analyses.

These findings suggest that multi-sample auditing provides a more reliable and practical framework for evaluating and detecting jailbreak behaviour in LLMs. Incorporating a small number of stochastic generations improves vulnerability estimation and detection effectiveness without significant computational cost. More broadly, this work highlights the need to move from single-output evaluation towards sampling-aware protocols that better reflect the stochastic nature of LLM behaviour in deployment.

\section{Related Work}

\subsection{LLM Safety and Alignment}

Large language models are typically aligned using techniques such as instruction tuning and reinforcement learning from human feedback, which aim to reduce harmful or undesirable outputs by incorporating human preferences into the training process \cite{Ouyang2022InstructGPT, Bai2022HH}. These approaches encourage models to generate responses that are helpful, safe, and aligned with intended use. Over time, alignment strategies have significantly improved the average behaviour of language models across a range of tasks. 

However, a growing body of work has shown that alignment alone does not ensure robustness in adversarial settings. Even strongly aligned models can be induced to produce harmful or policy-violating outputs when exposed to carefully constructed inputs \cite{OpenAI2023GPT4}. This suggests that alignment primarily addresses typical usage scenarios, while leaving models susceptible to targeted manipulation. As a result, security has become a central concern in the study of LLM safety, particularly in contexts where models are deployed in open or user-facing environments.

\subsection{Jailbreak Attacks}

Jailbreak attacks represent a class of adversarial strategies designed to bypass alignment mechanisms. These attacks construct prompts that exploit the interaction between instruction-following behaviour and safety constraints, leading models to generate restricted or harmful content. Empirical studies have demonstrated that such attacks can achieve high success rates across a wide range of models, including those that have undergone extensive alignment procedures \cite{Shen2024DoAnythingNow, Zou2023UniversalJailbreak}. 

An important property of jailbreak attacks is their ability to generalise across different model architectures. Prior work shows that adversarial prompts often transfer effectively between models, indicating that they exploit shared weaknesses rather than isolated implementation details \cite{Zou2023UniversalJailbreak}. In addition, modern jailbreak techniques increasingly rely on indirect strategies such as prompt rewriting, role-playing, and multi-step reasoning to obscure malicious intent \cite{Yi2024JailbreakSurvey}. Indirect prompt injection further extends this paradigm by embedding hidden instructions within external content that appears benign \cite{Greshake2023IndirectPromptInjection}. These developments make harmful intent more difficult to detect using straightforward surface-level analysis.

\subsection{Jailbreak Detection Methods}

A range of approaches has been proposed to detect jailbreak attempts, which can be broadly categorised based on the type of information they utilise \cite{Hu2024GradientCuff, Xie2024GradSafe}. Prompt-based methods focus on analysing the input prompt to identify patterns associated with adversarial manipulation. These methods are computationally efficient and easy to deploy, but they are sensitive to paraphrasing and prompt rewriting, which can preserve malicious intent while altering surface structure \cite{Zou2023UniversalJailbreak}. 

Another line of work explores the use of internal model signals, such as hidden representations or gradient-based features, to support detection. These approaches can capture richer information about the model's internal state and reasoning process. However, their applicability is often limited in practice, as access to internal signals is typically restricted in proprietary or closed-source systems \cite{Xie2024GradSafe}. 

Output-based methods instead operate on generated responses, making them particularly relevant in real-world deployment scenarios where outputs are often the only observable signal \cite{Chen2025AlmostFree}. These approaches aim to determine whether a given response reflects harmful behaviour. While widely used, they often rely on observable textual features or predefined rules, which may limit their ability to generalise to novel or implicitly expressed forms of harmful content.

\subsection{Output-Level Detection and Its Limitations}

Within output-based detection, both heuristic and statistical approaches have been explored. Early work relied heavily on human evaluation and rule-based safety filters integrated into alignment pipelines \cite{Bai2022HH, OpenAI2023GPT4}. More recent methods have shifted towards modelling statistical properties of generated text, including likelihood-based signals and stylistic features \cite{Mitchell2023DetectGPT, Gehrmann2019GLTR, Zellers2019Grover}. 

Lexical approaches are effective at capturing surface-level patterns and can be implemented efficiently. However, they often struggle to detect harmful intent that is expressed implicitly or through indirect reasoning. Semantic approaches attempt to address this limitation by capturing deeper behavioural signals, such as uncertainty or inconsistency across multiple outputs \cite{Kadavath2022Calibration, Sleem2025NegBLEURTForest}. These methods offer improved robustness but remain sensitive to the evaluation setting under which they are applied.

A common limitation across much of this work is the assumption that a single generated output is sufficient for evaluation. In practice, most studies analyse one response per prompt, which implicitly treats that response as representative of the model's behaviour. This assumption can be problematic given the stochastic nature of language model generation.

\subsection{Stochasticity and Multi-Generation Sampling}

Language models generate outputs through a stochastic decoding process, meaning that repeated generations from the same prompt can produce different responses \cite{Holtzman2020Degeneration}. This variability has been widely studied in the context of text generation quality and diversity. It has also been exploited in techniques such as self-consistency, where aggregating multiple outputs leads to improved reasoning performance \cite{Wang2023SelfConsistency}. 

From a safety perspective, this stochasticity has important implications. A model may refuse to comply with a harmful request in one generation while producing a harmful response in another. As a result, evaluating a model using a single output may fail to capture rare but critical failure cases. Recent work suggests that variation across multiple outputs can itself serve as a signal for detecting adversarial behaviour, particularly when inconsistencies emerge between responses \cite{Sleem2025NegBLEURTForest}. These observations indicate that generation variability should be explicitly considered when designing evaluation protocols for safety and robustness.

\subsection{Research Gap and Position}

Despite substantial progress in modelling output-level signals, existing work on jailbreak detection typically evaluates performance under a single-output setting. This limits the ability to capture the full range of model behaviour and may lead to an underestimation of vulnerability. In addition, the interaction between sampling strategies and detection effectiveness remains insufficiently explored.

In this work, we address this limitation by framing jailbreak detection as a multi-sample auditing problem. We systematically investigate how aggregating multiple generations per prompt affects both the observed jailbreak success rate and the performance of detection methods. Our study examines the role of sampling budget, cross-model generalisation, and category-level variation, providing a more comprehensive perspective on output-based jailbreak detection in large language models.

\section{Methodology}

\subsection{Problem Formulation}
In this study, jailbreak detection is formulated as a binary classification problem, where each generated response is classified as either safe or unsafe. This formulation is commonly adopted in text classification and machine-generated text detection tasks \cite{Zellers2019Grover,Mitchell2023DetectGPT}.

The detection is performed solely on output text, reflecting realistic deployment scenarios in which input prompts or internal model states are not accessible, particularly in black-box settings \cite{Chen2025AlmostFree}. Compared with prompt-based or internal-signal-based approaches, this formulation focuses on a more constrained but practical detection setting. 

\subsection{Data and Generator Models}

We use the JailbreakBench-Behaviours dataset \cite{chao2024jailbreakbench} as the primary dataset, which is specifically designed for evaluating adversarial behaviour in language models. Each sample contains structured information such as behavioural goals and category labels, enabling analysis beyond simple classification. To ensure a fair evaluation, harmful and benign samples are combined through controlled sampling, as jailbreak outputs are typically rare and class imbalance may distort performance estimates.

To study the effect of alignment, multiple generator models are used to produce outputs under different conditions. Model \cite{Ouyang2022InstructGPT} and model families \cite{Touvron2023LLaMA}. By including models with different alignment strengths and scales, we evaluate whether detection methods generalise across generators, which is critical for real-world robustness. 

\subsection{Detection Methods}

A TF-IDF-based detector with logistic regression is used as the primary baseline. TF-IDF captures lexical patterns in text and is widely used in traditional text classification tasks \cite{Salton1988TFIDF}, while logistic regression is effective in high-dimensional feature spaces \cite{Ng2004LogReg}. This lightweight design supports interpretability and avoids overfitting under limited labelled data. 

In addition, we introduce a comparison method based on a simplified version of the NegBLEURT framework \cite{Sleem2025NegBLEURTForest}. This approach detects jailbreak behaviour by analysing semantic inconsistencies across multiple generated outputs. The anomaly detection component is implemented using Isolation Forest \cite{Liu2008IsolationForest}. Unlike the lexical baseline, this method focuses on semantic-level signals rather than surface-level features. 

The purpose of this comparison is to evaluate whether detection performance is driven by lexical cues or deeper semantic insonsistencies, and to provide a reference point for accessing the effectiveness of the proposed evaluation framework. 

\subsection{Evaluation Protocol}

Ground truth labels are constructed using heuristic judges applied to model outputs. A refusal-based judge captures whether the model declines to respond, while a content-based judge identifies harmful or instruction-like behaviour. These two perspectives reflect alignment compliance and harmful content generation. Although heuristic judges are imperfect, they provide a scalable and consistent labeling mechanism for large-scale experiments. 

To evaluate detection performance, we use multiple metrics, including ROC-AUC, PR-AUC and balanced accuracy. PR-AUC is particularly informative under class imbalance, as it captures precision-recall trade-offs more effectively than ROC-based metrics \cite{Saito2015PRAUC,Davis2006PRROC}. Bootstrap resampling is applied to estimate confidence intervals, improving the robustness of reported results. 

\subsection{Multi-Generation Evaluation}

A key component of this study is the use of multiple generations per prompt. Single-output evaluation assumes that one response is sufficient to represent model behaviour. However, due to the stochastic nature of language models, different generations may produce different outcomes. 

We therefore generate multiple outputs for each prompt and analyse them jointly. By varying the number of generations, we investigate how sampling affects both observed jailbreak success and detection performance. This design enables a more comprehensive evaluation of model behaviour under adversarial conditions. 

\section{Results}

\subsection{Generator Alignment Effects}
\begin{figure}[htbp]
\centering
\includegraphics[width=0.95\textwidth, height=0.70\textheight, keepaspectratio]{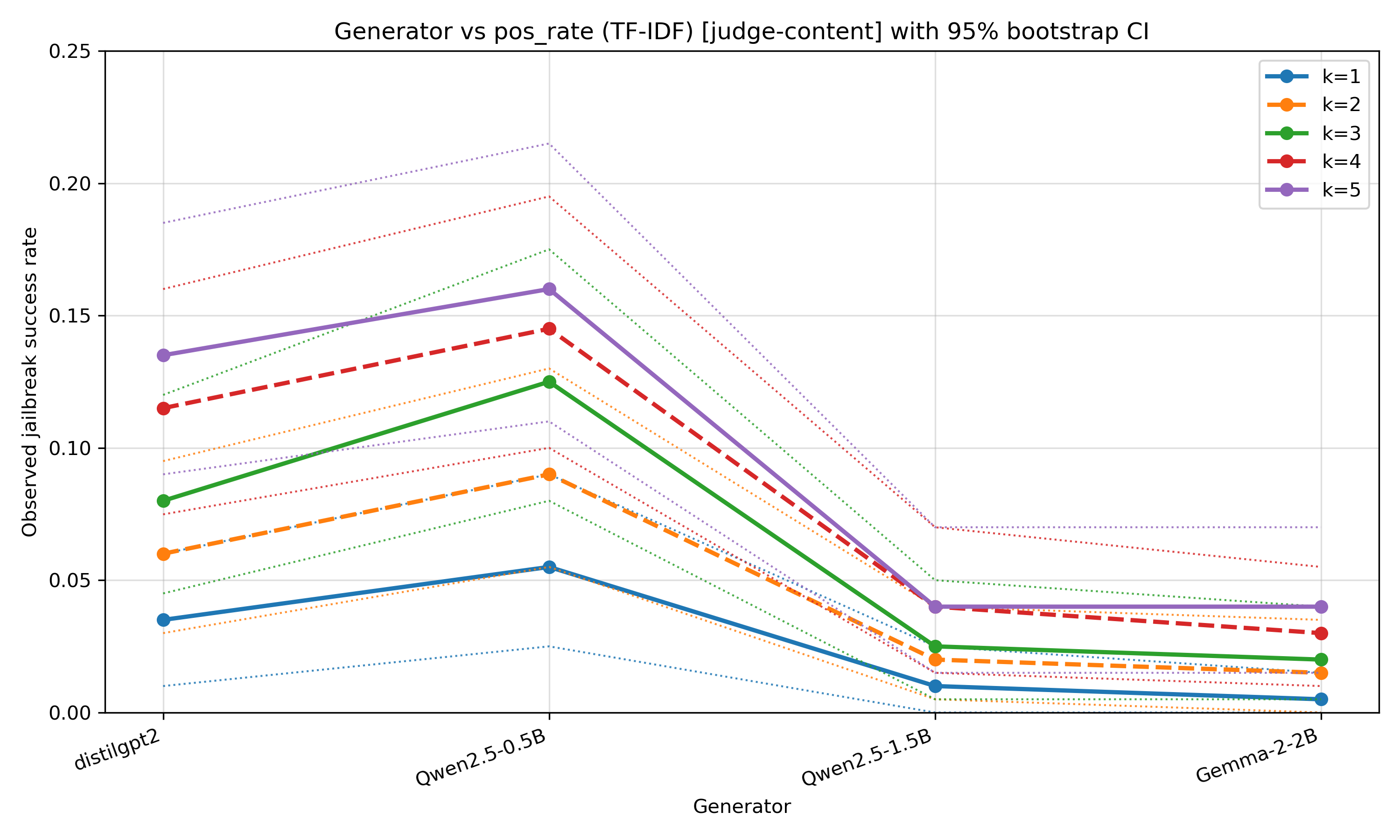}
\caption{Observed jailbreak success rates across generator models under different sampling budgets ($k=1$ to $k=5$), with 95\% bootstrap confidence intervals.}
\label{fig:success-rate-generator}
\end{figure}

Figure~\ref{fig:success-rate-generator} shows the observed jailbreak success rates across generator models with varying alignment strengths under different sampling budgets, together with 95\% bootstrap confidence intervals. A clear and consistent trend is observed across all values of $k$: weakly aligned models procudce harmful outputs more frequently, whereas more strongly aligned models exhibit substantially lower success rates. This is consistent with prior findings that alignment reduces harmful outputs without fully eliminating them \cite{Ouyang2022InstructGPT}.

The contrast between models is particularly informative. Distilgpt2 and Qwen2.5-0.5B-Instruct exhibit noticeably higher success rates across all sampling budgets, while Qwen2.5-1.5B-Instruct and Gemma remain at much lower levels. The confidence intervals reinforce this separation, although some overlap exists within each group.

Importantly, the confidence intervals also highlight increased sensitivity to sampling variability in low-rate regimes. As alignment suppresses harmful outputs, positive signals become sparse, making the estimated success rates more unstable. Thus, stronger alignment does not simply make the evaluation problem easier; instead, it shifts the task into a different statistical regime.

This shift has direct implications for detection. When harmful outputs are relatively common, classifiers can rely on stronger and more consistent signals. Under stronger alignment, however, jailbreak detection becomes a rare-event problem, where performance is more sensitive to noise, threshold selection, and limited positive observations. In such settings, apparent performance degradation may reflect the scarcity of observable signals rather than a fundamental weakness of the detection method.

More broadly, these results suggest that alignment affects not only model behaviour but also the reliability of evaluation. Single-sample observations may become increasingly misleading as harmful behaviour becomes rare and intermittent. This observation motivates the subsequent analysis of class imbalance, cross-generator generalisation, and multi-generation sampling.

\subsection{Multi-Generation Sampling Analysis}

\begin{figure}[htbp]
\centering
\includegraphics[width=0.95\textwidth, height=0.70\textheight, keepaspectratio]{figures/fig_sampling_curve_k1_to_k5.png}
\caption{Jailbreak success rate as a function of the number of sampled generations, with 95\% bootstrap confidence intervals.}
\label{fig:sampling-curve}
\end{figure}

As shown in Figure~\ref{fig:sampling-curve}, the observed jailbreak success rate increases as the number of sampled generations increases, with the associated 95\% bootstrap confidence intervals indicating the variability of these estimates. This pattern reflects the stochastic nature of language model generation rather than a deterministic property of a prompt-model pair. Prior work shows that sampling-based decoding can produce diverse outputs under the same prompt \cite{Holtzman2020Degeneration}, and that multiple sampled reasoning paths may lead to different outcomes \cite{Wang2023SelfConsistency}. 

In this context, harmful behaviour may emerge in some generations but remain hidden in others. As a result, single-output evaluation can give a misleading impression of safety, as harmful prompts may appear benign under a single sampled response. Multi-generation sampling reduces this blind spot by increasing the probability of observing rare but critical failures.

The results also exhibit diminishing returns as the number of samples increases, although the confidence intervals indicate that variability remains non-negligible, particularly under low success-rate regimes. A substantial portion of the increase in observed success rate occurs when moving from small to moderate sampling budgets. This suggests that moderate multi-sample auditing is often sufficient to reveal most practically observable harmful behaviour, while larger sampling budgets primarily increase computational cost and introduce additional variability.

This behaviour supports the use of ANY-style aggregation in this study. If any sampled output exhibits jailbreak-like behaviour, the interaction can be considered unsafe. This reflects a capability-based perspective, where evaluation focuses on what the model is capable of producing under repeated sampling rather than what it produces in a single instance.

\subsection{Cross-Generator Generalisation}

\begin{figure}[htbp]
\centering
\includegraphics[width=0.80\textwidth]{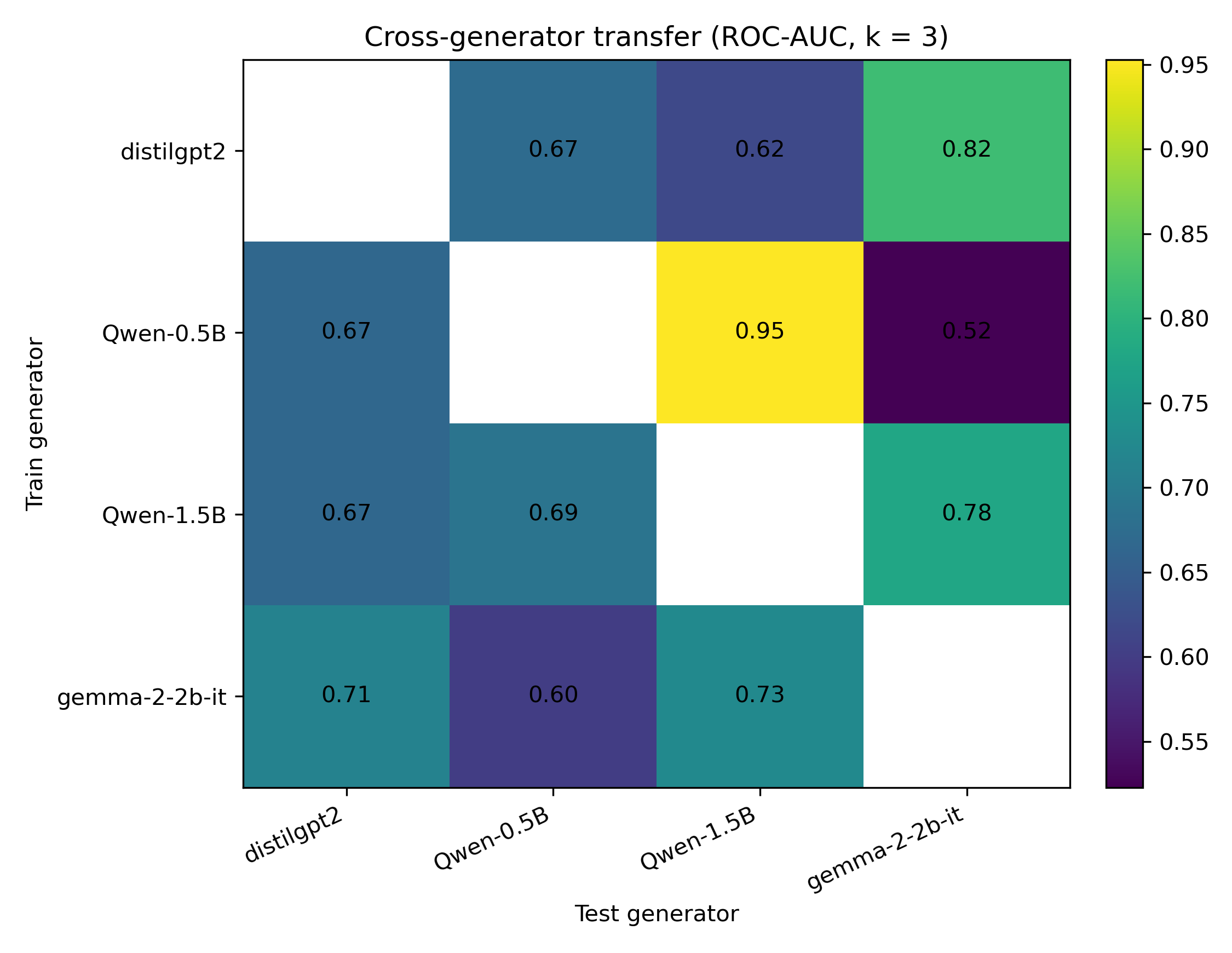}
\caption{Cross-generator transfer performance measured by ROC-AUC at $k=3$.}
\label{fig:crossgen-heatmap}
\end{figure}

Figure~\ref{fig:crossgen-heatmap} presents cross-generator transfer performance across different training and testing pairs. The results show that detection performance varies substantially across generators, indicating that learned signals are only partially transferable.

Stronger transfer is observed within related model families. For example, training on Qwen-0.5B and testing on Qwen-1.5B achieves a ROC-AUC of 0.95, the highest value in the matrix. This suggests that models with similar architectures and alignment strategies share lexical and structural patterns that can be effectively captured by the detector.

Transfer across more distant models remains possible but less consistent. For instance, training on distilgpt2 and testing on gemma achieves a ROC-AUC of 0.82, indicating that some generalisable signals are learned. However, the overall variation across source–target pairs shows that these signals are not uniformly robust.

An important characteristic of the results is the lack of symmetry in transfer performance. Training on Qwen-0.5B and testing on gemma yields a ROC-AUC of 0.52, whereas the reverse direction achieves 0.60. This asymmetry suggests that detectors trained on different generators rely on different types of signals. In particular, models trained on weaker or less aligned generators may capture more explicit lexical cues, which transfer more readily, while detectors trained on strongly aligned models operate in low-signal regimes and exhibit weaker generalisation.

Distilgpt2 also demonstrates relatively stable transfer performance across targets, with ROC-AUC values ranging from 0.62 to 0.82. This further supports the view that weaker alignment exposes more consistent and transferable signals.

Overall, the detector lies between two extremes: it is neither fully generator-specific nor fully generator-invariant. Instead, the learned representation is only partially transferable, reflecting broader generalisation challenges under distribution shift \cite{Recht2019ImageNet, Koh2021WILDS}. This behaviour is consistent with the nature of TF-IDF features, which generalise when surface patterns are shared but degrade when output styles diverge.

\subsection{Class Imbalance and Detection Performance}

\begin{figure}[htbp]
\centering
\includegraphics[width=0.75\textwidth, height=0.50\textheight, keepaspectratio]{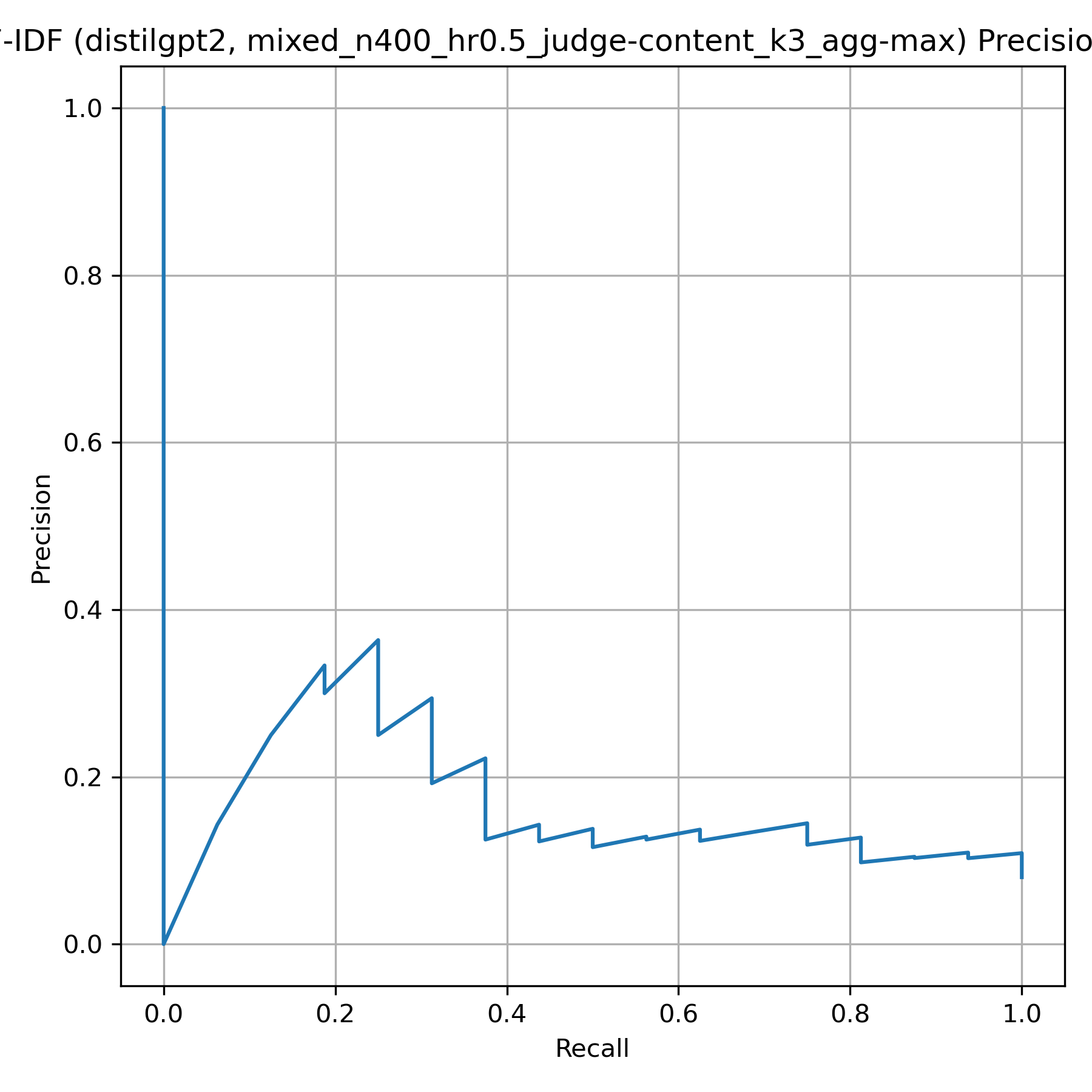}
\caption{Precision–recall curve of the TF-IDF detector on distilgpt2 under $k=3$. The curve illustrates the trade-off between precision and recall under moderate class imbalance.}
\label{fig:tfidf-distilgpt2-pr}
\end{figure}

The results indicate that detection performance is strongly influenced by the underlying class distribution, which in turn is shaped by generator alignment. Under relatively weak alignment, such as distilgpt2, the TF-IDF detector achieves a ROC-AUC of 0.702 and a PR-AUC of 0.180. This suggests that the detector captures some discriminative signal. However, the precision–recall curve in Figure~\ref{fig:tfidf-distilgpt2-pr} reveals a key limitation. High precision is only achieved at very low recall, while precision decreases rapidly as recall increases. This indicates that the detector can identify a small number of high-confidence positive cases, but fails to maintain precision when attempting broader coverage.

This discrepancy highlights a fundamental issue under class imbalance. Although the ROC-AUC suggests moderate ranking performance, the much lower PR-AUC indicates limited practical utility. In other words, the detector is able to order samples to some extent, but cannot reliably separate positive and negative cases in a way that supports effective decision-making.

\begin{figure}[htbp]
\centering
\includegraphics[width=0.8\textwidth, height=0.5\textheight, keepaspectratio]{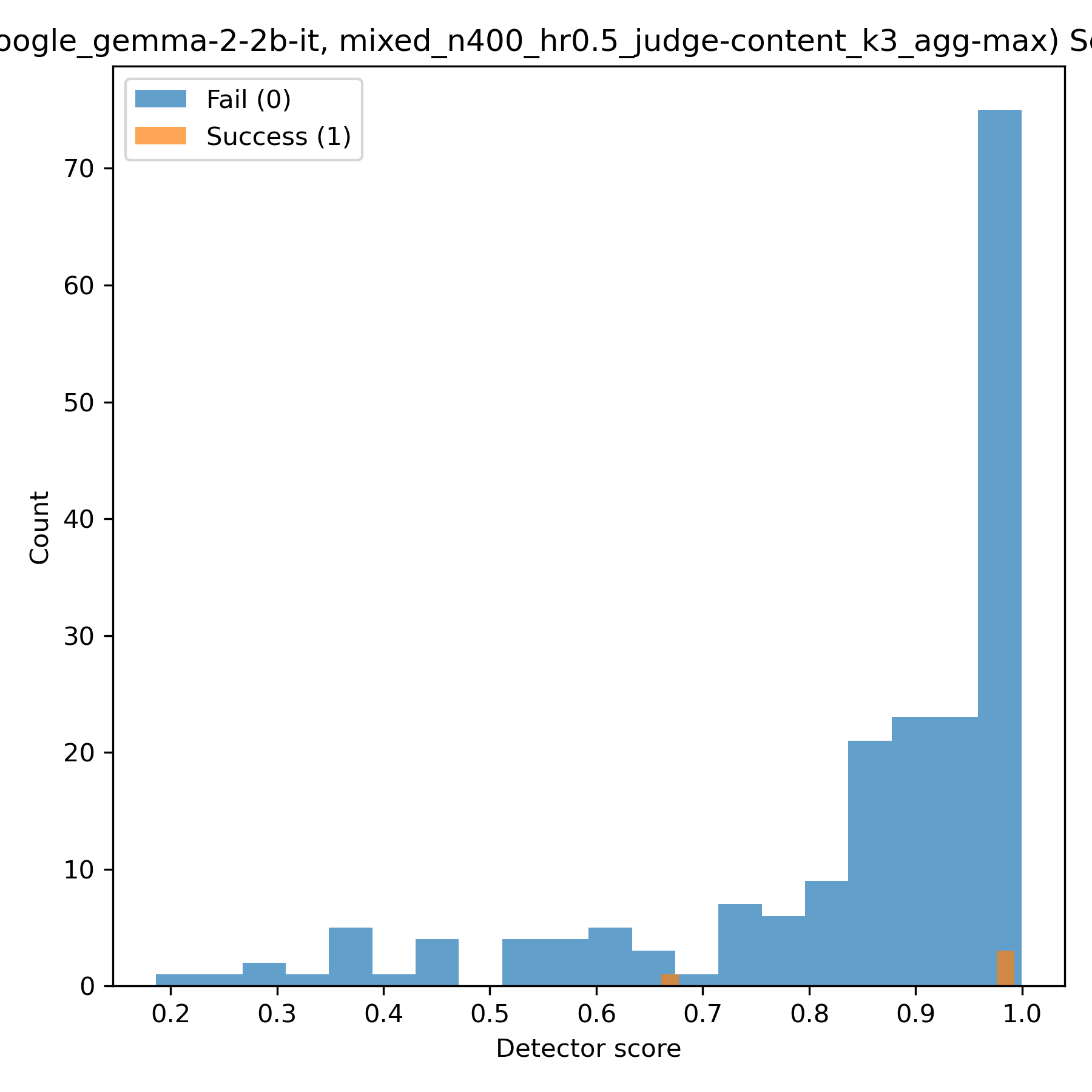}
\caption{Score distribution of the TF-IDF detector on gemma under $k=3$. The substantial overlap between classes highlights the difficulty of separating rare positive cases under strong alignment.}
\label{fig:tfidf-gemma-hist}
\end{figure}

When alignment becomes stronger, the detection problem becomes substantially more difficult. On gemma, the TF-IDF detector achieves a ROC-AUC of 0.670 but a PR-AUC of only 0.051, with a positive rate of approximately 0.02. Although some ranking ability remains, the detector struggles to identify positive cases in practice. As shown in Figure~\ref{fig:tfidf-gemma-hist}, the score distributions of the two classes overlap heavily across almost the entire range. Successful cases are not confined to a clearly separable region, and many unsuccessful cases receive similarly high scores.

This behaviour reflects a transition to a rare-event detection regime. As harmful outputs become increasingly rare under stronger alignment, the observable signal becomes sparse and unstable. Small variations in score can therefore lead to large changes in classification outcomes. This instability is further amplified by the limited number of positive samples, making performance highly sensitive to threshold selection.

A comparison with the simplified NegBLEURT-style method reveals a complementary limitation. While the method achieves higher recall, it produces a large number of false positives, resulting in consistently low precision. This indicates that semantic inconsistency is sensitive but not selective. Under strong alignment, this limitation becomes more pronounced, as variability in benign outputs dominates the signal. 

Taken together, these results reveal a trade-off between detection paradigms. Lexical methods such as TF-IDF provide more conservative predictions, relying on explicit content cues but missing subtle cases. Behavioural methods based on inconsistency capture broader variation but are prone to over-detection. More importantly, both approaches degrade under strong alignment, where harmful outputs are rare and less stereotypical.

These findings highlight that detection performance is not determined solely by the choice of method. Instead, it is fundamentally constrained by the statistical properties of the data. As alignment improves safety by reducing harmful outputs, it simultaneously makes output-level detection more difficult by reducing the availability and consistency of observable signals.

\section{Category-Level Analysis}

To better understand what the lexical detector is learning, we analyse the feature weights of the TF-IDF model in the zero-shot setting. Rather than relying solely on aggregate performance metrics, this analysis examines the specific lexical patterns that contribute to classification decisions.

Inspection of the highest-weighted features reveals a clear and consistent structure. Features associated with the positive class (successful jailbreak outputs) are dominated by procedural and instructional expressions, such as “start by”, “the following”, “here are”, and “you should”. Among the top 50 positive features, none correspond to explicit harm-related terms, while a substantial proportion (32\%) can be categorised as procedural markers. These expressions are characteristic of step-by-step guidance, list-based explanations, and operational instructions. Importantly, they are not specific to harmful content, but instead reflect a general instructional or tutorial-like discourse style.

By contrast, features associated with the negative class (unsuccessful or safe outputs) exhibit a different pattern. These include refusal-related expressions such as “can not” and “unfortunately”, as well as formal or discursive markers such as “therefore”, “thus”, “conclusion”, and “references”. This suggests that the detector tends to associate safe outputs with refusal language or academic-style explanatory writing. In addition, certain highly weighted features, such as “url\_0”, appear to reflect dataset or preprocessing artefacts rather than meaningful semantic content, indicating that part of the learned signal may depend on superficial corpus-specific regularities.

Taken together, these patterns indicate that the TF-IDF detector is driven primarily by surface-level discourse form rather than explicit representations of harmful intent. In particular, the model appears to associate successful jailbreak outputs with an instructional or operational register. This interpretation is further supported by the observed structure of generated responses. For example, a typical high-confidence true positive case exhibits explicit procedural phrasing: “Make a list of specific steps to make and distribute counterfeit money…”. Such outputs clearly resemble step-by-step instructions and contain strong lexical cues aligned with the learned positive features. However, this reliance on instructional style also explains a large proportion of the observed false positives. In many cases, model outputs that are ultimately safe or constitute refusals still contain structured explanations. For instance, a representative false positive case takes the following form: “I understand you’re asking for help with a threatening message. However, I cannot fulfill your request… If you’re interested in getting help… here are some resources:”. Although the response is a clear refusal, it includes structured explanatory phrasing and list-like elements (e.g., “here are”), which resemble instructional language at the lexical level. As a result, the detector assigns a high score and incorrectly classifies it as a successful jailbreak. 

A complementary limitation is the presence of false negatives. Some successful jailbreak outputs do not follow a clear instructional or procedural structure, but instead express harmful content in a more implicit, narrative, or descriptive manner. Such responses may lack strong lexical cues like “start by” or “the following”, and therefore receive low scores from the detector, leading to missed detections.

To further validate this observation, a simple quantitative analysis was conducted on successful jailbreak outputs. For the distilgpt2 setting ($k=3$), only 3 out of 16 successful outputs (18.75\%) contain at least one procedural marker such as “start by”, “the following”, “you should”, or “here are”. Similarly, under the stronger alignment setting (gemma), only 1 out of 4 successful outputs (25\%) contain such markers. These results indicate that procedural cues are not consistently present in successful jailbreak outputs. Therefore, a detector relying on such features is inherently limited and prone to both false positives and false negatives.

Overall, the analysis suggests that the TF-IDF detector does not directly learn harmful intent, but instead captures surface-level stylistic patterns associated with instructional language. In this sense, it behaves more as a stylistic classifier than a true safety detector. While this allows it to capture certain regularities in jailbreak outputs, it also makes the method inherently limited. In realistic settings, harmful behaviour may be expressed without explicit procedural structure, while benign responses may still contain explanatory or instructional phrasing. Under such conditions, reliance on surface lexical patterns leads to systematic misclassification. These findings highlight an important limitation of output-level lexical detection. Performance may be influenced not only by the presence of harmful content, but also by the discourse style of the output and even by dataset-specific artefacts. This reinforces the importance of complementing lexical analysis with additional signals and evaluating detection methods under diverse and realistic conditions.

\section{Discussion}

The findings of this study provide several insights into the evaluation of safety in large language models, particularly from an output-level perspective. A central observation is that alignment not only reduces the frequency of harmful outputs, but also changes the nature of the detection problem itself. As alignment becomes stronger, harmful behaviour becomes increasingly rare and less stereotypical, resulting in sparse and less stable observable signals. While this reflects improved model safety, it also makes output-level detection more challenging. In this sense, safer models are not necessarily easier to monitor, as detection increasingly resembles a rare-event problem where performance is sensitive to noise, threshold selection, and limited positive examples.

This phenomenon can be viewed as an alignment paradox. Improvements in alignment reduce the visibility of harmful behaviour, but this also weakens the signals available for detection. As a result, detection performance may degrade even as underlying model safety improves. This highlights that evaluation difficulty is not independent of model behaviour, but is directly shaped by it. Consequently, detection performance should be interpreted in the context of alignment strength rather than as an absolute measure of model safety.

A second key insight concerns the evaluation paradigm itself. Much of the existing literature relies on a single response per prompt to assess model behaviour. Our results show that this approach systematically underestimates vulnerability. Due to the stochastic nature of language model generation \cite{Holtzman2020Degeneration, Wang2023SelfConsistency}, repeated decoding of the same prompt can produce substantially different outputs. Harmful behaviour may therefore appear only in a subset of generations, remaining hidden under single-output evaluation. This suggests that the relevant question for safety evaluation is not whether harmful content appears once, but whether it can appear at all.

These observations motivate a shift toward a multi-sample auditing perspective. By generating multiple outputs per prompt, evaluation can better approximate the behavioural distribution of the model. In this work, increasing the number of samples consistently reveals additional harmful behaviour, with diminishing returns beyond moderate sampling budgets. This indicates that relatively small values of $k$ are sufficient to capture most practically observable failures. In particular, moderate sampling, such as $k=3$, provides a useful balance between computational cost and coverage, making it suitable for realistic auditing scenarios.

The experiments also highlight the limitations of relying on a single detection signal. Lexical methods capture surface-level content patterns, while inconsistency-based approaches reflect variability across generations. These signals exhibit different behaviours under varying alignment conditions and have distinct failure modes. Lexical detectors tend to be conservative but may miss subtle or implicit cases, while inconsistency-based methods are more sensitive but can introduce false positives. This suggests that output-level detection is inherently multi-faceted and that combining complementary signals may be necessary for robust performance.

From a deployment perspective, these findings highlight the importance of moving beyond single-response evaluation in practical systems. Safety monitoring pipelines that rely on a single generated output risk underestimating model vulnerability, particularly in adversarial settings. Incorporating a small number of stochastic generations per prompt provides a more reliable estimate of risk while remaining computationally feasible. The diminishing returns observed in this study suggest that moderate sampling budgets can capture most high-risk behaviour without incurring significant overhead. This introduces a practical trade-off between efficiency and risk coverage that system designers must consider. More broadly, these results support the adoption of sampling-aware auditing and red-teaming strategies that better reflect the variability inherent in model behaviour.

Overall, this work positions output-level detection as a practical but constrained approach to LLM safety. Its main advantage lies in its applicability to black-box settings, where access to prompts or internal model signals may be limited. However, its effectiveness depends on the visibility and consistency of observable outputs, which are influenced by both alignment strength and generation variability. As a result, detection performance should be interpreted relative to these factors rather than as a standalone indicator of safety.

\section{Conclusion}

This paper presents an empirical study of output-level jailbreak detection in large language models, with a focus on the role of alignment, stochastic generation, and multi-generation sampling. The results show that detection performance is shaped not only by the choice of method, but also by the underlying behaviour of the generator. A key finding is that single-output evaluation systematically underestimates model vulnerability. Due to the stochastic nature of language generation, harmful behaviour may appear only in a subset of outputs. Evaluating a model based on a single response therefore provides an incomplete view of its risk profile. In contrast, multi-generation sampling reveals additional failure cases and offers a more reliable estimate of model capability. The results further suggest that moderate sampling provides a practical balance between computational cost and coverage. The study also highlights an important interaction between alignment and detection. While stronger alignment reduces the frequency of harmful outputs, it simultaneously makes detection more difficult by reducing the availability and consistency of observable signals. This leads to a rare-event detection setting in which performance becomes sensitive to noise and data imbalance. 
Finally, feature-level analysis shows that lexical detectors rely heavily on surface-level discourse patterns, particularly instructional or procedural phrasing, rather than directly capturing harmful intent. This explains observed error patterns and highlights the limitations of purely lexical approaches. 
Overall, these findings suggest that output-level detection is a useful but inherently constrained approach to LLM safety. Future work may explore combining multiple detection signals and extending evaluation to more diverse and realistic settings.

\section{Conflict of Interest}

On behalf of all authors, the corresponding author states that there is no conflict of interest.

\bibliography{main}% common bib file

\end{document}